%% file: 00_paper.tex
\documentclass[conference]{IEEEtran}
\IEEEoverridecommandlockouts
\usepackage{balance}       
\usepackage{booktabs}
\usepackage{enumitem}
\usepackage{color}
\usepackage{xspace}
\usepackage{cite}
\usepackage{amsmath,amssymb,amsfonts}
\usepackage{algorithmic}
\usepackage{graphicx}
\usepackage{textcomp}
\usepackage{xcolor}
\usepackage[draft, bookmarks=false]{hyperref}

\makeatletter
\def\ps@IEEEtitlepagestyle{%
  \def\@oddfoot{\mycopyrightnotice}%
  \def\@oddhead{\hbox{}\@IEEEheaderstyle\leftmark\hfil\thepage}\relax
  \def\@evenhead{\@IEEEheaderstyle\thepage\hfil\leftmark\hbox{}}\relax
  \def\@evenfoot{}%
}
\def\mycopyrightnotice{%
  \begin{minipage}{\textwidth}
  \centering \scriptsize
  Copyright~\copyright~2021 IEEE. Personal use of this material is permitted. Permission from IEEE must be obtained for all other uses, in any current or future media, including\\reprinting/republishing this material for advertising or promotional purposes, creating new collective works, for resale or redistribution to servers or lists, or reuse of any copyrighted component of this work in other works by sending a request to pubs-permissions@ieee.org.
  \end{minipage}
}
\makeatother

\def\BibTeX{{\rm B\kern-.05em{\sc i\kern-.025em b}\kern-.08em
    T\kern-.1667em\lower.7ex\hbox{E}\kern-.125emX}}
\begin{document}

\title{Graph Neural Networks to Predict Sports Outcomes\\
}

\author{\IEEEauthorblockN{Peter Xenopoulos}
\IEEEauthorblockA{New York University\\
New York, NY\\
\href{mailto:xenopoulos@nyu.edu}{xenopoulos@nyu.edu}}
\and
\IEEEauthorblockN{Claudio Silva}
\IEEEauthorblockA{New York University\\
New York, NY\\
\href{mailto:csilva@nyu.edu}{csilva@nyu.edu}}}

\IEEEoverridecommandlockouts
\IEEEpubid{\makebox[\columnwidth]{978-1-7281-6251-5/20/\$31.00 \textcopyright 2021 IEEE \hfill} \hspace{\columnsep}\makebox[\columnwidth]{ }}

\maketitle
\IEEEpubidadjcol

\newcommand{\peter}[1]{{\color{blue} Peter: [{#1}]}}
\newcommand{\claudio}[1]{{\color{red} Claudio: [{#1}]}}

\begin{abstract}
\input{00_abstract}
\end{abstract}


\input{01_intro}
\input{02_relatedwork}
\input{03_methods}
\input{04_results}
\input{05_discussion}
\input{06_conclusion}

\section*{Acknowledgment}
\input{07_acks}

\balance

\bibliographystyle{IEEEtran}
\bibliography{00_ref}

\end{document}

%% file: 00_abstract.tex
Predicting outcomes in sports is important for teams, leagues, bettors, media, and fans. Given the growing amount of player tracking data, sports analytics models are increasingly utilizing spatially-derived features built upon player tracking data. However, player-specific information, such as location, cannot readily be included as features themselves, since common modeling techniques rely on vector input. Accordingly, spatially-derived features are commonly constructed in relation to anchor objects, such as the distance to a ball or goal, through global feature aggregations, or via role-assignment schemes, where players are designated a distinct role in the game. In doing so, we sacrifice inter-player and local relationships in favor of global ones. To address this issue, we introduce a sport-agnostic graph-based representation of game states. We then use our proposed graph representation as input to graph neural networks to predict sports outcomes. Our approach preserves permutation invariance and allows for flexible player interaction weights. We demonstrate how our method provides statistically significant improvements over the state of the art for prediction tasks in both American football and esports, reducing test set loss by 9\% and 20\%, respectively. Additionally, we show how our model can be used to answer ``what if'' questions in sports and to visualize relationships between players.

%% file: 01_intro.tex
\section{Introduction}
Sports analytics has been gaining increased traction as detailed spatio-temporal player event and trajectory data become readily available, enabled by enhanced data acquisiton systems~\cite{Assuncao19SportsAnalytics}. To acquire these data, practitioners rely on methods that range from manual annotation~\cite{Liu13Opta} to automated tracking systems that record player locations at high frequencies. Oftentimes, the collected sports data are large, complex, and heterogeneous, which presents interesting and unique research directions in information retrieval, visualization, but in particular, machine learning-based prediction~\cite{Brefeld17SISports}. 

The increase in high-fidelity sports data streams has been a boon for many prediction-based problems in sports, such as valuing players in soccer~\cite{Decroos19VAEP, Yurko2020LSTM}, detecting high leverage moments in esports~\cite{Xenopoulos2020CSGOWPA}, or assessing decision making in basketball~\cite{cervone2014pointwise}. The way in which the state of a game is represented is central to aforementioned sports analytics efforts. Game states can contain both the \textit{global} game context, such as each team's score or the game time remaining, as well as \textit{local} context like player locations stored in X/Y coordinates. Typically, a game state has been represented as a single vector, since a vector is a straightforward abstraction and readily applicable for many out-of-the-box data mining techniques. 

As spatio-temporal data, such as player actions and trajectories, become more prevalent, game state vectors are also increasingly including spatially-derived features. When constructing spatially-derived features, practitioners must enforce permutation invariance with respect to the ordering of players, since they often use a single vector as input. Consequently, to create permutation invariant spatially-derived features, practitioners often turn to one of (1) a permutation invariant function, like the average, to aggregate player-specific information into global features~\cite{Xenopoulos2020CSGOWPA, Yang17ResultPred}, (2) a strict ordering of players through role assignment, such as assigning each player to a distinct game role~\cite{Makarov17PredictingWinner, Lucey13PlayerRoles, Sha17TreeAlignment}, or (3) keypoint-based feature construction, where features are ordered in relation to an important location on the field~\cite{Mehrasa2018Deep, Sicilia19MicroActions, Yurko2020LSTM}. Example keypoints include the ball-carrier in American football or the goal in soccer. Using this construction, features can then be defined as the closest player distance to the goal in soccer, second closest player distance to the ball, and so on.

While the aforementioned methods are oftentimes readily interpretable, they discard information in a player's local neighborhood or require identification of a keypoint before preprocessing the data in order to anchor spatial information. Furthermore, although keypoints are usually identifiable in ball-based sports such as soccer or basketball, they are ambiguous for sports such as esports, or competitive video gaming, where no ball exists. To address these issues, we introduce a sport-agnostic, graph-based framework to represent game states. In our framework, a game state is represented by a fully-connected graph, where players constitute nodes. Edges between players can represent the distance between the players or be assigned a constant value. Using our graph representation of game states, we present permutation invariant graph neural networks to predict sports outcomes. We demonstrate our method's efficacy over traditional vector-based representations for prediction tasks in American football and esports. Our contributions are (1) a sports-agnostic graph representation of game states to facilitate prediction in sports, (2) a graph-based neural network architecture to predict outcomes in American football and esports, and (3) case-studies that not only validate our models by identifying known relationships in sports but also demonstrate how our modeling approach enables ``what if'' analysis in sports. 

The rest of the paper is structured as follows. In Section~\ref{sec:related_work}, we review relevant literature on feature construction for sports prediction and on graphs in sports analytics. Section~\ref{sec:methods} presents our graph-based representation of game states, graph neural networks, and the model architectures we use. In Sections~\ref{sec:results} and \ref{sec:discussion}, we describe our results and present use cases of our method. In Section~\ref{sec:conclusion}, we conclude the paper.

%% file: 02_relatedwork.tex
\section{Related Work} \label{sec:related_work}

\subsection{Feature Construction}
Feature construction is a fundamental aspect of sports analytics. With an influx of spatio-temporal data, game states are increasingly including sport-specific spatial features, such as closest player distance to the goal in soccer or average player distance to the ball-carrier in American football. To craft these features, practitioners must identify a keypoint in the game. For example, to estimate the success of soccer passes Power~et~al.~\cite{Power17Passing} calculate the speed, distance and angle of the nearest defender to the passer. In this instance, the keypoint is the passer. Ruiz~et~al.~\cite{Ruiz17Leicester} present an expected goals model for soccer, where they use angle and distance to goal, making the defending team's goal the keypoint. Decroos~et~al.~\cite{Decroos19VAEP} build a scoring likelihood model where they include features such as distance and angle to goal, as well as distance covered during a soccer action.

We can also order players, and then use their features as input, if the ordering given any permutation of the players in a game is invariant. In Mehrasa~et~al.~\cite{Mehrasa2018Deep}, the authors propose a permutation invariant sorting scheme based on an anchor object. The first player in the sorting scheme is the anchor player. Then, the next player is the one closest to the anchor player, and the next player is the one second-closest to the anchor player, and so on. Similarly, Sicilia~et~al.~\cite{Sicilia19MicroActions} order player coordinate features based on distance to the scoring basket. Similarly, Yurko~et~al.~\cite{Yurko2020LSTM} create features for each player based on their relative location and direction compared to the ball-carrier in American football. These features are then ordered based their distance to the ball-carrier.

Another way to enforce permutation invariance on game state features is through role assignment. Using a role assignment scheme, features can then be tied to a particular position, e.g., $Forward_x$, $Forward_y$, and so on. Role assignment can work well in sports which maintain a rigid structure. For example, Makarov~et~al. \cite{Makarov17PredictingWinner} use historical stats to predict a player's role in Defense of the Ancients (DOTA) 2, a popular esport where players generally play very specialized roles. Using these predicted roles, they then assign players in a team so that each role is occupied by a player. Lucey~et~al.~\cite{Lucey13PlayerRoles} introduce a role assignment method where the current player formation is compared against a formation template. Sha~et~al.~\cite{Sha17TreeAlignment} propose a tree-based method to estimate player roles which goes through a hierarchical approach of assigning roles based on a template, and then partitioning the data. 

Lastly, one way to ensure permutation invariance is through global aggregations. Aggregations are common in esports, due to a lack of keypoints. For example, Xenopoulos~et~al.~\cite{Xenopoulos2020CSGOWPA} aggregate team equipment values and minimum distances to bombsites for each game state in Counter-Strike. Yang~et~al.~\cite{Yang17ResultPred} aggregate each team's gold, experience, and deaths in DOTA to estimate win probability throughout the game. Makarov~et~al.~\cite{Makarov17PredictingWinner} consider the total number of healthy and damaged players, rankings of players, and total number of grenades remaining for each team to predict round winners in Counter-Strike.

\subsection{Graphs in Sports}
Graphs have found limited application in sports, with past works focusing primarily on network analysis or multi-agent trajectory prediction. Concerning network analysis, Passos~et~al.~\cite{Passos11Networks} use graphs, where nodes are players and edges represent number of passes between the players, to represent passing networks in water polo. The authors then used this information to differentiate between successful and unsuccessful patterns of play. Gudmundsson and Horton~\cite{Gudmundsson17Spatiotemporal} provide a summary of network analysis in sports, along with a comprehensive review of spatio-temporal analysis in sports.

Graphs have also proven useful for multi-agent modeling in sports. Sports situations can be represented with fully-connected graphs using nodes as players. Kipf~et~al.~\cite{KipfFWWZ18} introduce a variational autoencoder model in which the encoder predicts the player interactions given the trajectories, and the decoder learns the dynamical model of the agents. Recently, more works have focused on attention based mechanisms. Hoshen~et~al.~\cite{Hoshen17} introduce an attentional architecture to predict player movement that scales linearly with the number of agents. Yeh~et~al.~\cite{Yeh19MultiAgentGen} propose a graph variational recurrent neural network model to predict player trajectories. Ding~et~al.~\cite{Ding20Attention} use a graph-based attention model to learn the dependency between agents and a temporal convolution network to support long trajectory histories. While understanding how players move is an important focus on sports analytics, predicting outcomes, such as who will win the game, has broad and established applications for teams assessing player value or for bookmakers setting odds. Our work combines the traditional sports analytics objective of predicting sports outcomes with the graph-based representation prevalent in multi-agent sports trajectory modeling.

%% file: 03_methods.tex
\section{Graphs for Sports} \label{sec:methods}

\subsection{Game State Graph Construction}
\label{sec:graph_construction}
Let $S_t$ be a game state which contains the complete game information at time $t$. For example, $S_t$ can contain global information, such as team scores and time remaining, as well as player-specific information like location, velocity or other player metadata. Accordingly, $S_t$ is represented by the set $\{z_t, X_t\}$, where $z_t$ is a vector of global game state information and $X_t = \{x_t^1, x_t^2, ..., x_t^N\}$ is a set containing all player information, where $x_t^p$ is a vector containing the information for player $p$. $N$ is the number of players. 

Since $X_t$ is an unordered set, some consistent ordering is required to build a model which returns stable output across different permutations of the same input. As previously discussed in section~\ref{sec:related_work}, practitioners typically perform some operation on $X_t$, such as taking the average of its features or ordering its elements, to construct permutation invariant features. However, we seek a representation that avoids upfront global feature aggregation or role assignment. Furthermore, we want to apply a permutation invariant function to such a representation. To do so, we can represent game states as graphs, and use them as inputs to permutation invariant graph neural networks.

\begin{figure}
    \centering
    \includegraphics[width=\linewidth]{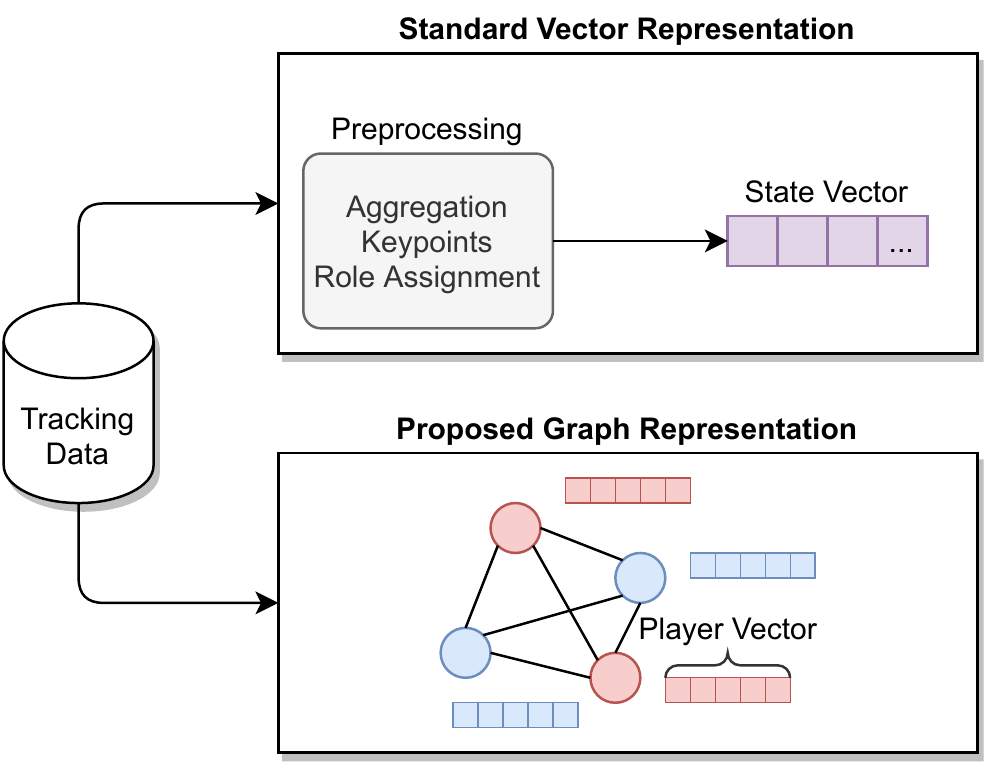}
    \caption{Our game state representation transforms raw player tracking data into a fully-connected graph where each player is described by a node feature vector, compared to the standard approach of representing a game state as a vector where features are constructed using keypoints, role-assignment, or aggregations. Thus, our method does not require any problem-specific preprocessing steps.}
    \label{fig:graph_construction}
\end{figure}

Consider a graph $G_t = (V_t, E_t)$, where $V_t$ represents the nodes and $E_t$ the edges. We use $G_t$'s nodes to represent player information of game state $S_t$. Setting $V_t = X_t$, we ensure that each player is a node in $G_t$ with an associated feature vector $x^p_t$. To ensure that every player is connected to one another, we let $e_{ij} = 1, \forall i,j$. This representation includes self-loops for every node. An alternate representation is to set edge weight $e_{ij}$ to the distance between player $i$ and player $j$. $e_{ij}$ does not necessarily need to be symmetric. For example, distances in esports are often asymmetric, as shown in~\cite{Xenopoulos2020CSGOWPA}. We show our game state and graph construction process in Figure~\ref{fig:graph_construction}. Our representation avoids any sport or problem-specific preprocessing steps, like aggregations, keypoint-definitions or role assignment, since we only need to take tracking data and transform it into a graph.

\subsection{Graph Attention Networks}
\label{sec:gat}
Using $G_t$ as input, we can apply graph neural networks to learn game outcomes from a given game state. Graph convolutional networks (GCNs)~\cite{Kipf17} and graph attention networks (GATs)~\cite{VelickovicCCRLB18} are popular graph neural network architectures. Both architectures accept a set of nodes as input, $\{h_1, h_2, ..., h_N\}, h_i \in \mathbb{R}^F$, where $N$ represents the number of nodes in the graph, and $F$ represents the number of features for each node. The output of both architectures is a set of node representations $\{h_1^{'}, h_2^{'}, ..., h_N^{'}\}, h_i^{'} \in \mathbb{R}^{K}$.

We first parameterize a weight matrix $\mathbf{W} \in \mathbb{R}^{F \times K}$. To produce a set of transformed nodes, for each node we calculate

\begin{equation} \label{eq:node_rep}
    h_i^{'} = \sigma \Bigg( \sum_{j\in \mathcal{N}_i} e_{ij} h_j \mathbf{W} \Bigg)
\end{equation}

\noindent where $\sigma$ is an activation function, $\mathcal{N}_i$ represents the neighborhood of node $i$, and $e_{ij}$ represents the edge weight from node $j$ to node $i$. Effectively, a node is transformed as the weighted average of its neighbors, where the weights are produced from a function learned during model training. The node representations can be fed into another GAT layer or into a pooling layer which can downsample the graph by producing coarse representations. The entire set of node representations can also be pooled through a global average pool, whereby we produce a vector containing the average values of the $K$ features of each element $h_i^{'}$. Such an operation is permutation invariant.

The main difference between GCNs and GATs are that GATs estimate attention coefficients between neighboring nodes. To do so, we use a feedforward network parameterized by a weight vector $\mathbf{a} \in \mathbb{R}^{2 K}$. The attention coefficient $e_{ij}$, which estimates the importance of node $j$'s features to node $i$, is calculated as 

\begin{equation} \label{eq:attn}
    e_{ij} = \frac{ \mathrm{exp}(\mathrm{LeakyReLU}(\mathbf{a}^T [h_i \mathbf{W} \| h_j  \mathbf{W} ]))}{\sum_{k \in \mathcal{N}_i } \mathrm{exp}(\mathrm{LeakyReLU}(\mathbf{a}^T [h_i \mathbf{W} \| h_k  \mathbf{W} ]))}
\end{equation}

\noindent where $\|$ represents a concatenation. 

Velickovic~et~al.~\cite{VelickovicCCRLB18} note that multi-head attention can be useful for stabilizing the self-attention process. To do so, one specifies $K$ independent attention mechanisms and performs the same transformation as in Equation~\ref{eq:node_rep}. We can create transformed node representations $h_i^{'}$ by doing

\begin{equation} \label{eq:node_rep}
    h_i^{'} = \Bigg\|_{k=1}^K \sigma \Bigg( \sum_{j\in \mathcal{N}_i} e_{ij}^k h_j \mathbf{W} \Bigg)
\end{equation}

\noindent where $e_{ij}^k$ is the attention weight from the $k$-th attention mechanism. To implement GCNs and GATs and average pooling, we use the Spektral library in Python~\cite{Spektral}.

\subsection{Candidate Models}
\label{sec:models}
To test if graph neural networks improve upon traditional sports analytics methodology, we consider a variety of models. Since a single feature vector is commonly used to represent sports game states, we structure our models in the same fashion, where we take a game state $G_t$ with associated outcome $Y_t$, and want to estimate some function $f$ where $f(G_t) = \widehat{Y_t}$. Our baseline model, referred to as the ``state model'', is based on such a representation. The input to our baseline model is a game state vector that contains global features, along with aggregated player features. Such a baseline model is common in sports analytics, and is used in~\cite{Yurko2020LSTM, Yang17ResultPred}.

\begin{figure}
    \centering
    \includegraphics[scale=0.69]{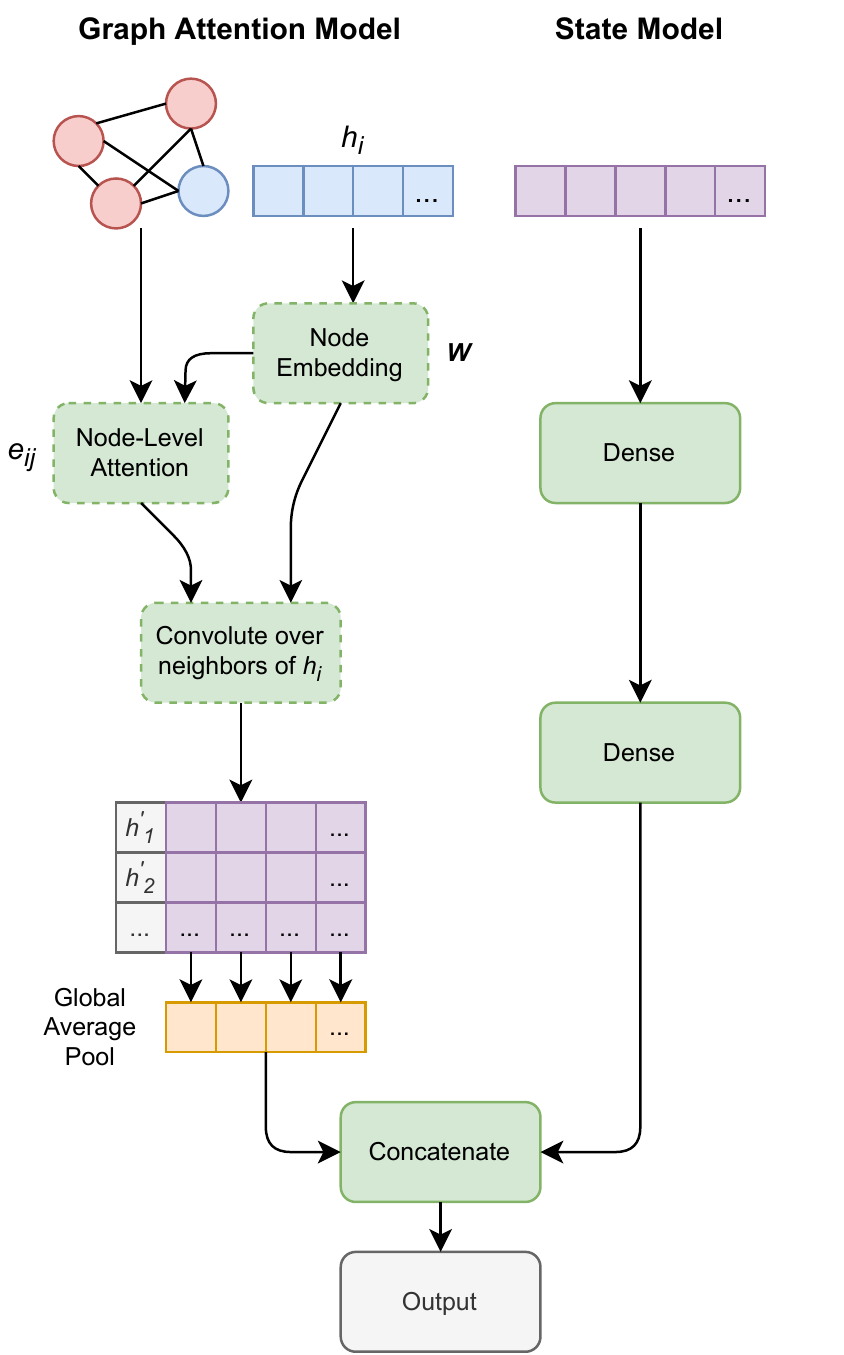}
    \caption{The GAT branch of the combined GAT+State model considers a game state graph as input and outputs transformed node representations using estimated attention weights $e_{ij}$. These new node representations are then average-pooled. The state model branch is a simple feedforward network. The outputs of the GAT and state branches are concatenated and passed through a final dense layer to predict the sports outcome of interest.}
    \label{fig:gat_model}
\end{figure}

We consider two graph-based models to compare against our baseline. The first is a graph attention network, as described in Section~\ref{sec:gat}. The second is a graph convolutional network (GCN)~\cite{Kipf17}. In our GCN model, nodes are transformed as shown in equation~\ref{eq:node_rep}, however $e_{ij}$ are \textit{not} estimated as shown in equation~\ref{eq:attn}. Instead, $e_{ij}$ are set to the inverse of the distance between player $i$ and player $j$. We use the inverse of the distance since, if node $j$ is closer to node $i$, it will have a higher weight when computing~\ref{eq:node_rep}. Intuitively, we would expect proximal nodes to have more influence over each other. The graph input to the GCN architecture is still fully-connected. When using GAT-based models, we use one attention head. We explore the effects of varying the attention head parameter in Section~\ref{sec:attn_heads}. Finally, we also consider an architecture which takes both a game state's graph and vector representations as inputs. We visualize this architecture in Figure~\ref{fig:gat_model}. In total, we consider five models: a state model, a GAT-only model, a GCN-only model, a GAT+State model, and a GCN + State model.

\subsection{Sports Prediction Tasks}
We investigate graph-based game state representations across prediction tasks in two different sports. We focus on American football as it is a team and ball-based sport, much like soccer or basketball, as well as Counter-Strike: Global Offensive, a popular esport. 

\subsubsection{Predicting Yards Gained in American Football}
Offensive American football plays typically involve passing or running the ball to advance it towards the defending team's endzone. When a player attempts to advance by running with the ball, the play is referred to as a ``rushing'' play. Predicting rushing yards gained given the current alignment of players is a common task in sports analytics for American football. For example, Yurko~et~al.~\cite{Yurko2020LSTM} used yards gained as the target for their ball-carrier model. Using this model, they then valued ball-carriers on their yards gained above expectation. The problem formulation of predicting yards gained in American football is a regression task where we estimate $Y_t \in \mathbb{R}$, which represents the amount of yards in the horizontal direction the ball-carrier gains until the play ends, given a single game state $G_t$. We refer to this task as the ``NFL task.''

Our baseline game state vectors for the NFL task contain the down, distance to go, ball-carrier velocity and displacement, and 11 features containing the distances to the $i$-th closest defender to the ball-carrier. In the graph representation, every player node is described by a player's X/Y coordinates, velocity, displacement since the last state, difference in location and speed to the ball-carrier, average distance to other players, and flags for if the player is on the offensive team and if he is the ball-carrier.

\subsubsection{Estimating Win Probability in Esports}
Estimating win probability is a fundamental task in sports analytics, with particular importance to sports betting and player valuation. In sports betting, win probability models guide odds for live betting. For player valuation, many systems revolve around valuing players based on how their actions change a team's chance of winning or scoring~\cite{Xenopoulos2020CSGOWPA, Decroos19VAEP}. In this prediction task, we aim to predict round winners in the popular esport, Counter-Strike: Global Offensive (CSGO), a five-on-five shooter where two sides, denoted T and CT, seek to eliminate each other and either aim to plant a bomb as the T side, or defuse the bomb (if planted) as the CT side. We refer to this task as the ``CSGO task.''

In a CSGO match, two teams play best-of-30 rounds on a ``map', which is a virtual world. Each team plays the first 15 rounds as one of the T or CT sides, and then switches starting at the 16th round. At the beginning of every round, each side spends money earned in previous rounds to buy equipment and weapons. To win a round, the T side attempts to plant a bomb at one of two bombsites, and the CT side's goal is to defuse the bomb. Each side can also win a round by eliminating all members of the opposing side. When a player dies, they can no longer make an impact in the round.

Consider a game state $G_t^r$ which occurs in round $r$ at time $t$. Round $r$ is associated with an outcome, $Y_r$, which is equal to 1 if the CT side wins the round, and 0 if the CT side loses. The resulting prediction task is to estimate $P(Y_r = 1 | G_t^r)$. The formulation of predicting win probability in CSGO given a single game state is consistent with past win probability estimation in esports literature~\cite{Xenopoulos2020CSGOWPA, Yang17ResultPred, Makarov17PredictingWinner}. 

Our baseline game state vectors for the CSGO task contain the time, starting equipment values, current equipment values, total HP and armor, and scores for each team. Every player node is described by a player's X/Y/Z coordinates, health and armor remaining, total equipment value and grenades remaining, distance to both bombsites and indicators for player side, if the player is alive, if the player has a helmet and if the player has a defuse kit, and a one hot encoding of the part of the map the player is located.

\subsubsection{Assessing Performance} \label{sec:assessing_performance}
We assess model performance using mean squared error and mean absolute error for the NFL task (regression), and log loss and AUC for the CSGO task (classification). We use 70\% of game states for training, 10\% for model validation, and 20\% for testing in both of our prediction tasks. For CSGO, we create a different model for each of the six maps, which are different virtual worlds, since each map is different, unlike the standardized playing surface in most conventional sports. 


%% file: 04_results.tex
\section{Results} \label{sec:results}

\subsection{Data}
\label{sec:data}
For our NFL task, we use Next Gen Stats tracking data, which contains player and ball locations, from the first six weeks of the 2017 National Football League (NFL) season. The data was publicly released for the duration of the 2018 NFL Big Data Bowl. To create our data set of rushing plays, we remove plays with penalties, fumbles, and those involving special teams, like kickoffs and punts. We define a rushing play as starting when the rusher receives the handoff, and ending when the rusher is tackled, called out of bounds, or scores a touchdown. In total, we used 4,038 plays containing 129,859 game states.

For our classification task in CSGO, we use the csgo package from Xenopoulos~et~al.~\cite{Xenopoulos2020CSGOWPA} to parse 7,222 professional local area network (LAN) games from January 1st, 2017 to January 31st, 2020. To acquire the CSGO data, we downloaded game replay files (also known as a demofile) from a popular CSGO website, HLTV.org, which hosts competitive CSGO replays. We use LAN games since the environment is more controlled, as players are physically in the same space, and every player has low latency with the game server. A single game can contain performances on multiple maps, and each map performance is contained in a single demofile. We parsed 9,414 demofiles into JSON format at a rate of one game state per second. In total, we used 14,291,069 game states.

\subsection{Model Results}

\begin{figure*}
    \centering
    \includegraphics[scale=0.6]{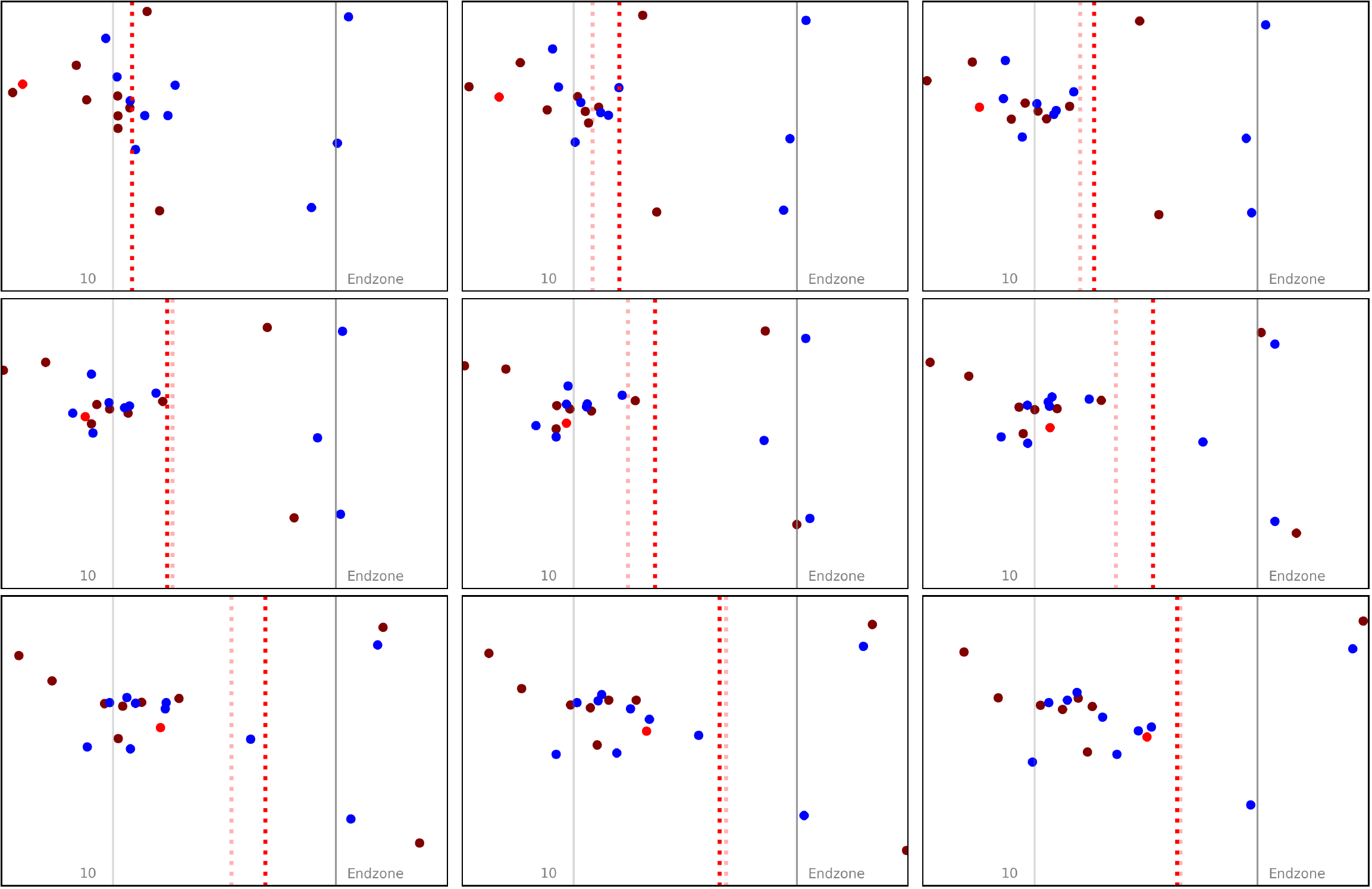}
    \caption{Left-to-Right, Top-to-Bottom, a sequence of game states from an NFL rushing play. The rusher is red, offensive players are brown, and defending players are blue. As the rusher moves into the unoccupied space at the bottom of the field, his expected ending yard line (red dashed line) intuitively increases from over five yards compared to the prediction at the beginning of the play. The faint red dashed line is the previous game states's expected ending yard line. We see the states which provide large yard gains increases over previous states are those where the rusher finds large amounts of uncontested space ahead of him.}
    \label{fig:football_sequence}
\end{figure*}

\begin{table}[]

\caption{Graph neural network performance on prediction tasks for American football (NFL) and esports (CSGO). The combined GCN+State model performs best on the NFL task, and the GAT+State model, as described in Figure~\ref{fig:gat_model}, performed best on the CSGO task.}

\centering
\begin{tabular}{@{}rcccc@{}}
\multicolumn{1}{c}{} & \multicolumn{2}{c}{\textit{NFL}} & \multicolumn{2}{c}{\textit{CSGO}} \\ \toprule
\multicolumn{1}{c}{\textbf{Model}} & \textbf{MSE} & \textbf{MAE} & \textbf{Log Loss} & \textbf{AUC} \\ \midrule
\textit{State} & 35.39 & 3.05 & 0.4351 & 0.8715 \\
\textit{GCN} & 33.40 & 3.02 & 0.6111 & 0.7399 \\
\textit{GAT} & 33.64 & 2.73 & 0.4465 & 0.8684 \\
\textit{GCN + State} & \textbf{30.62} & \textbf{2.68} & 0.4357 & 0.8718 \\ 
\textit{GAT + State} & 33.58 & 2.93 & \textbf{0.4276} & \textbf{0.8753} \\ \bottomrule
\end{tabular}

\label{tab:results}
\end{table}

In this section, we compare our candidate models quantitatively and qualitatively, not only by analyzing the performance metrics described in Section~\ref{sec:assessing_performance}, but also by confirming known relationships in American football and CSGO. Table~\ref{tab:results} presents the results of our model testing procedure. Overall, we see that graph architectures outperformed the more classical state-vector models. We also see that in some instances, predefined weights using player distances work better than weights estimated using GATs, and vice versa, depending on the sport. Finally, we also see that our graph-based models encode known relationships in both American football and CSGO.

For the NFL task, due to the small amount of NFL plays in our training set, we repeat the train/validation/test procedure outlined in \ref{sec:models} 30 times, and we report the mean model performance for MSE and MAE in Table~\ref{tab:results}. We perform paired t-tests between each model and the baseline state model, where the null hypothesis is defined as $H_0: MSE_{state} - MSE_{model} = 0$ and the alternate hypothesis is defined as $H_a: MSE_{state} - MSE_{model} > 0$. There are significant differences between all of the models' MSE's and the state model's MSE, with the GCN + State model showing the largest test statistic ($t = 9.95$), and the GAT + State model showing the smallest ($2.71$). Of note, the state model only performed better in one trial compared to the GCN + State model. We show statistically significant improvements of 9\% over the state of the art single-state input models as described in Yurko~et~al.~\cite{Yurko2020LSTM}. They also describe a sequence-based model that uses multiple game states to predict rusher end yard position which performs better than any single-state model. However, the feature construction in~\cite{Yurko2020LSTM} still relies on ordering and anchor objects, and we discuss in the next section how our approach may yield more interpretable results.

The results from the NFL task suggest that using inter-player distances in a GCN, when paired with state vector information, may be beneficial over estimating player weights through a GAT-based model. The GCN took twice as long to train as the state model, and the GAT took almost three times as long as the state model. Interestingly, we also created a GAT model for the NFL task which only included the ball-carrier and the defensive players. The aforementioned model outperformed the state model with an MSE of 34.32, which was significantly lower, however it did not outperform any of the other candidate models.

For CSGO, we report the map-weighted performance mean, as some maps are more common in our data than others. Since we had millions of game states, we used one train/validation/split rather than the 30 trials procedure for the NFL task, since there was considerably less sampling variability. In Table~\ref{tab:results}, we see that the combined GAT + State model outperformed the state model. Our solution improves upon the state of the art for the CSGO task in Xenopoulos~et~al.~\cite{Xenopoulos2020CSGOWPA} by 20\%. Additionally, the GAT model performed well in the CSGO task compared to the GCN model, indicating that estimating attention coefficients greatly aided in predicting CSGO round winners as opposed to using player distances. One of the main differences is that the GATs were able to correctly assign attention coefficients of near-zero to dead players. This behavior is intuitive, since dead players are unable to make an impact for the rest of a round. 

We can also assess our proposed models for how well they capture well-known sports phenomena. For the following evaluation, we consider the GAT-only model, however, the other graph-based models also confirm the same phenomena. The first phenomena we consider is the idea that American football rushers will often seek open parts of the field in order to advance the ball for longer before being stopped by the defending team. The intuition here is that the larger the unoccupied space ahead of the rusher, the greater distance we would expect him to run. In Figure~\ref{fig:football_sequence}, we see a rushing play that starts near the 10 yard line, which is close to the defending team's endzone, meaning the team in red/orange has a scoring opportunity. We can observe the rusher (red) exploiting the gap in the defensive line (blue players) to advance, which intuitively increases his expected ending yard line. In fact, in the middle row of Figure~\ref{fig:football_sequence}, we see a large increase in the rusher's expected ending yard line, particularly in the last frame of this row, when the rusher clears the gap and is furthest away from defenders. 

Win probability, the quantity we estimate in the CSGO task, is commonly used in player valuation models. Xenopoulos~et~al.~\cite{Xenopoulos2020CSGOWPA} valued player actions by the change in win probability that those actions incurred. One finding of the study was that kills are one of the most important actions in CSGO, as they greatly change a team's win probability. One of the downsides to the approach in~\cite{Xenopoulos2020CSGOWPA} was that beyond damage and kill events, the model produced relatively flat win probabilities for most of a round. Because of this problem, events that positioned a player to achieve a kill, such as movement across the map, or engagements that resulted in little damage (such as those displacing a player), were not explicitly modeled, and thus were difficult to value. 

\begin{figure}
    \centering
    \includegraphics[scale=0.55]{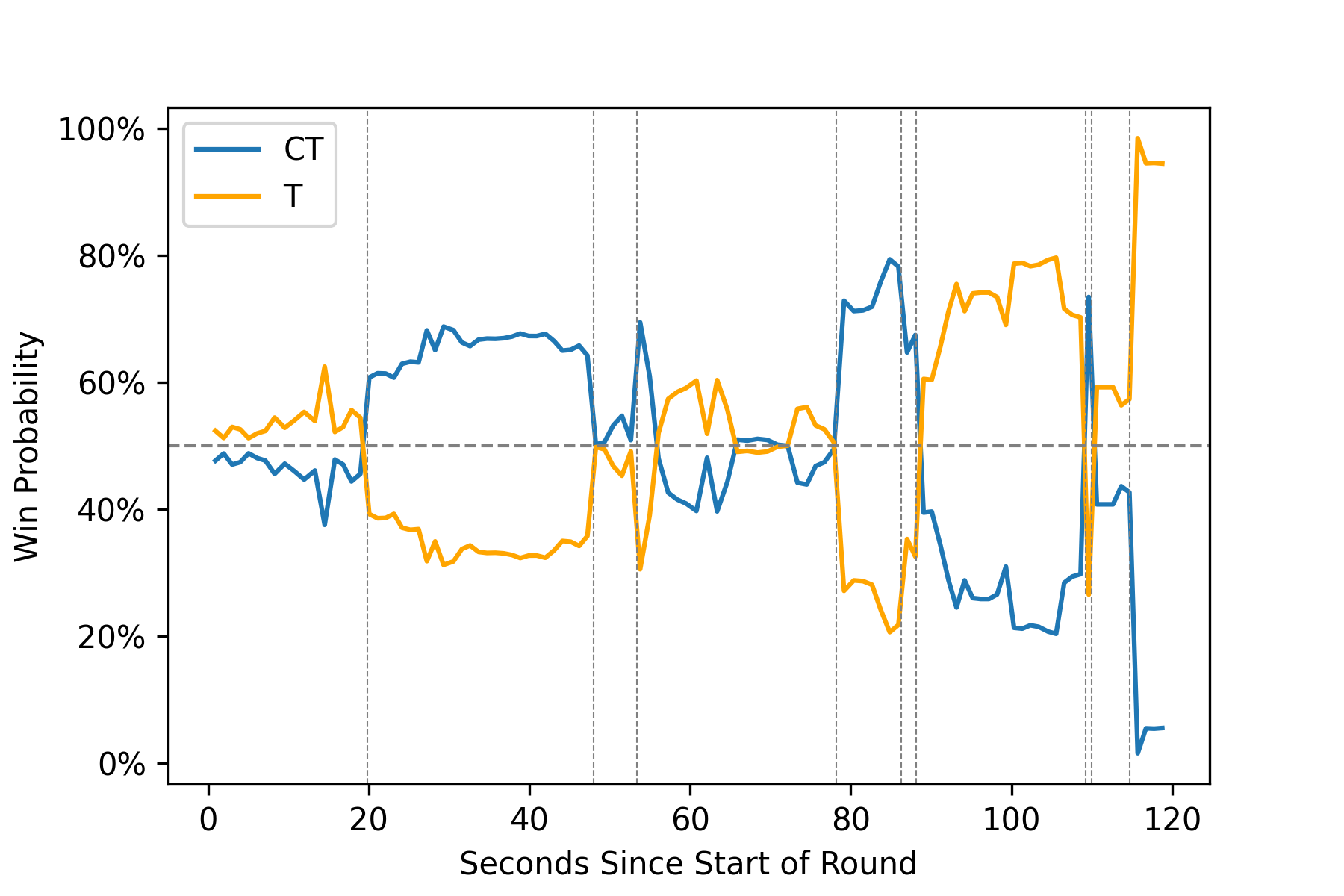}
    \caption{Our CSGO model closely tracks influential game events, such as kills (gray vertical lines), and improves upon previous CSGO win probability models, which often had significant portions of rounds with unchanging predictions due to inability to account for player-specific positions and equipment. Thus, actions which do not result in a kill or damage, such as player movement, can now be valued.}
    \label{fig:csgo_win_probability}
\end{figure}

Using our GAT model, we estimate win probability over the course of a CSGO round for both the CT and T sides in Figure~\ref{fig:csgo_win_probability}. Consistent with prior findings, we see that large jumps in win probability are due to kills (gray vertical lines). Additionally, we see far more variance in the win probability plots over the course of the round than in~\cite{Xenopoulos2020CSGOWPA}, which is due to our model more explicitly considering individual player locations, their equipment and their relationship with other players. 

\subsection{Enabling ``What If'' Analysis in Sports}
Understanding how model predictions change due to input perturbations is important for practitioners in sports analytics such as game analysts, coaches, players, or even curious fans. In both information symmetric games like American football, where players know the locations of all other players, and asymmetric games like CSGO, where players only initially know the locations of their teammates, decision making is fundamentally rooted in an understanding of players in the game space. One of the drawbacks to current keypoint-based models is that changes in player location only change the prediction if the relationship to the keypoint is changed. Accordingly, any type of ``what if'' analysis can be unsatisfactory to stakeholders since small changes in player location will do little to change model output. However, we know that relationships exist not only between a player and a keypoint, but also between other players. Since our graph construction ensures that every node is connected to one another, changes in one player's characteristics can propagate to all other player nodes. 

\begin{figure}
    \centering
    \includegraphics[scale=0.45]{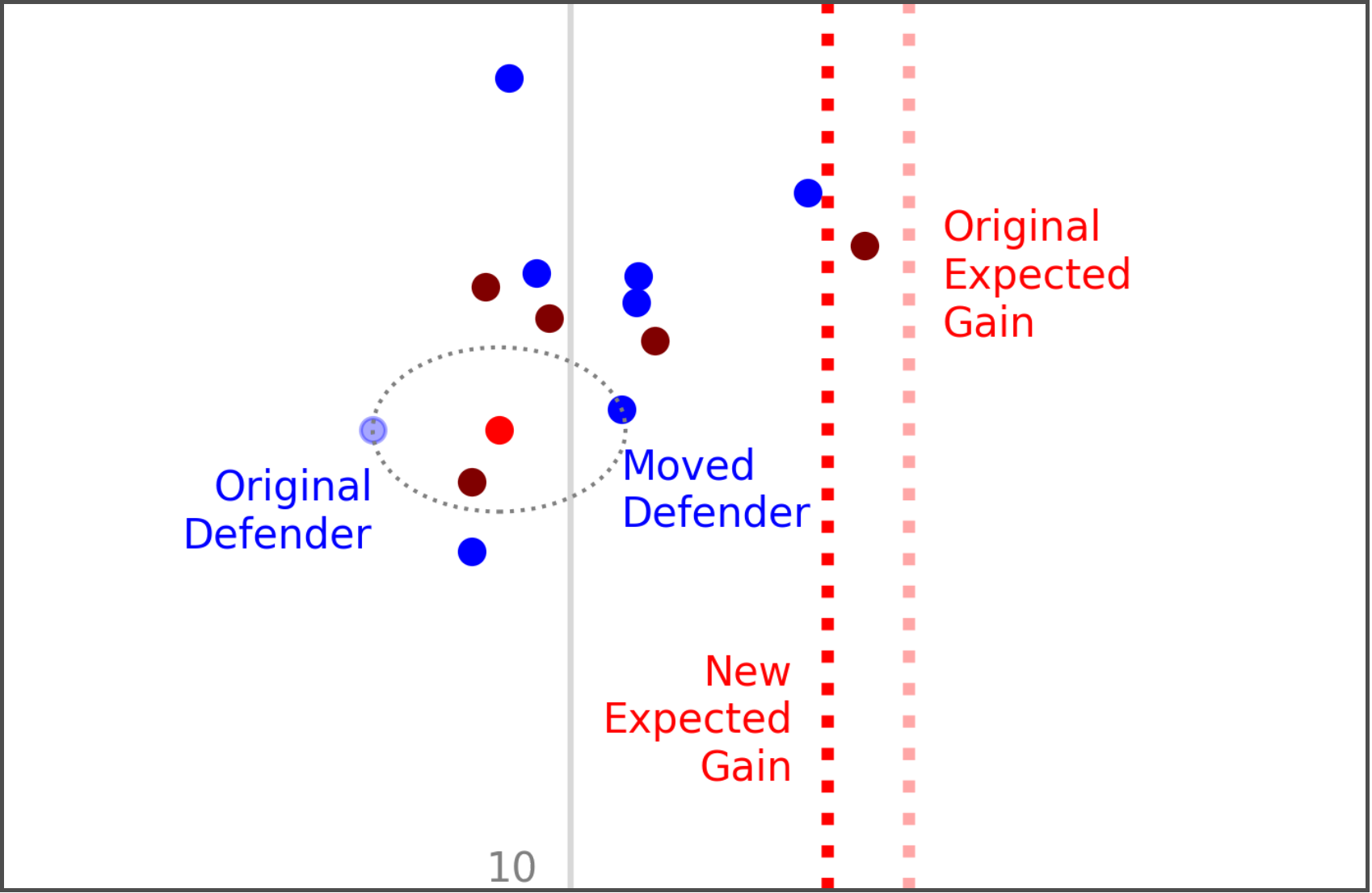}
    \caption{Our GAT model is able to consider how a player's relationship to other players, as opposed to a keypoint, changes the prediction. Moving the original defender along a circle with fixed radius to the rusher, a state-vector model would show no difference, but our GAT model intuitively predicts the rusher has a lower expected gain (red dotted line) than the gain with the defender in the original position (faint red dotted line).}
    \label{fig:football_what_if}
\end{figure}

To illustrate the above issue, consider a model that takes a game state vector as input, like the one described in Section~\ref{sec:models}, which takes 11 features that denote the $i$-th defender distance to the rusher in American football. If we move a defender along a fixed radius circle to the rusher, the state-vector model output will be unchanged. However, we know that the positioning of the defender is incredibly important. For example, if the defender is behind the rusher, it is unlikely that the defender will continue to significantly contribute to the play. On the other hand, if the defender is in the rusher's running path, he has greater potential to impede the rusher's advancement. We would expect in the latter situation, if our model reflected real relationships, that our predicted rusher gain would be smaller. Our model confirms this relationship, which we show in Figure~\ref{fig:football_what_if}. Since our defender's distance to the rusher is unchanged, the change in rusher expected gain is purely from the moved defender's changing relationship with other players.

Using model output, coaches and analysts can begin to value the impact of player decisions and portray the implications to players. Consider a scenario where the original defender position in Figure~\ref{fig:football_what_if} reflects a missed tackle by that defender. One way to value the implication of the tackle is to use the outcome of the play. We know that outcomes such as touchdowns or the rusher's gain may be informative, but also reflect the entirety of the play and can be noisy. Thus, without the GAT model, we may only be able to convey to the player that the implication of his missed tackle is that the rusher will advance. Such an insight is obvious. However, by using the GAT model, we are able to tell the player from Figure~\ref{fig:football_what_if} that the missed tackle caused the rusher's predicted gain to increase from an average of 3.54 yards (solid red line) to 4.28 yards (faint red line), which represents roughly a 20\% increase. With this insight, a player may better be able to consider the implication of the missed tackle.

In CSGO, the CT side defends two bombsites to prevent the T side from planting the bomb. Accordingly, players often assume static positions and coordinate among themselves on how they will defend a bombsite. Prior CSGO win probability models did not explicitly update win probability based on player position, and the main determinants of win probability were players remaining, their health and their equipment. In Figure~\ref{fig:csgo_win_probability}, we see that although this remains true when using a GAT to predict CSGO win probability, there exists significant change in win probability between kills. These changes are driven primarily by player movement. Thus, it becomes possible for a CSGO analyst, coach or player to assess his or her position, which enables ``what if'' analysis for CSGO.

\begin{figure}
    \centering
    \includegraphics[scale=0.45]{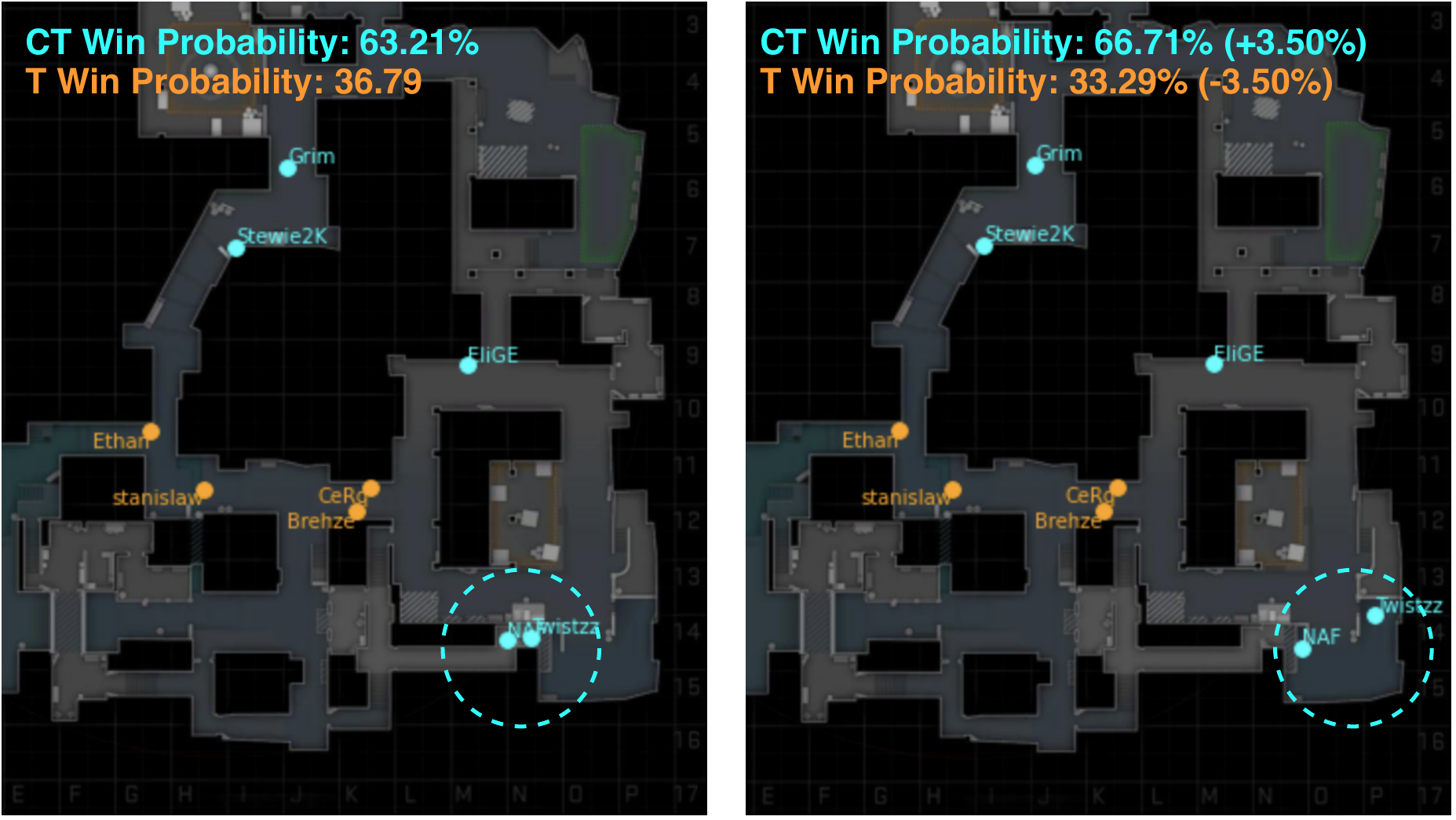}
    \caption{Our GAT model can be used to enhance game analysis and coaching. Positioning is an important aspect to CSGO gameplay. Here, we see that using just a slightly different positioning, the CT side can increase their probability of winning the round by 3.5\%. }
    \label{fig:csgo_what_if}
\end{figure}

Consider the five CT versus four T scenario posed in Figure~\ref{fig:csgo_what_if}. Here the two CT players in the bottom right of both frames are defending bombsite ``A''. On the left frame, which represents the original game state, both players are located side by side in a building that does not overlook the bombsite. We can manually exploring different positional setups by moving players around, and search for setups with increase the CT side's win probability. We proposed the showed the situation in Figure~\ref{fig:csgo_what_if} to a coach and an analyst from two top CSGO teams (both teams were in the world top 30). The feedback we received was that the increase in win probability was intuitive, as the players set up a ``crossfire'', which is when two view directions create a perpendicular angle. Additionally, the analyst remarked that the players in the second scenario have ``deeper map information'', as their lines of sight are larger and non-overlapping, as compared to the first scenario.

In practice, our model can drive a visual analytics system that allows one to position players on a playing surface or change player attributes, such as health or equipment, to answer questions such as ``how much does our chance of winning increase if player $A$ stood at spot $X$ with weapon $W$ as opposed to spot $B$''. These interactive systems are well received by the sports analytics community. For example, in the Chalkboarding system by Sha~et~al.~\cite{Sha18CHI}, users were able to engage in interactive analytics to discover most advantageous scenarios to make a successful shot in basketball.

%% file: 05_discussion.tex
\section{Discussion} \label{sec:discussion}

\begin{figure*}
    \centering
    \includegraphics[scale=0.63]{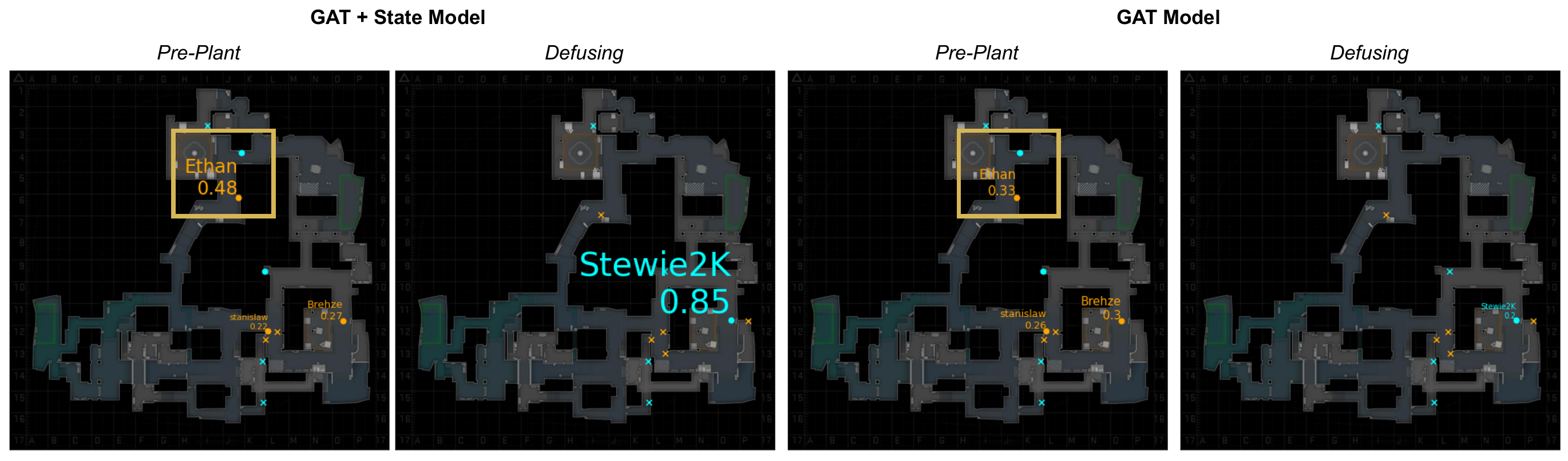}
    \caption{Estimated attention coefficients vary across the different models. The GAT-only model is unable to account for the bomb plant and low time remaining in the ``defusing'' scenario. Additionally, we see that in the GAT+State model, the players in the yellow bounded box have much higher average attention
    coefficients, potentially indicating that their engagement is high leverage.}
    \label{fig:csgo_attn}
\end{figure*}

\subsection{Visualizing Attention Coefficients}
One benefit of using GATs to predict sports outcomes is a byproduct of the training process -- the attention coefficients described in Section~\ref{sec:gat}. We observed that the attention coefficients generated from the CSGO task reflect game phenomena. For example, if a player was dead (and thus, could not create an impact for the rest of the round), the estimated average attention coefficient between said player and other players was effectively near zero. However, interestingly, we observed that on average, T sided players had much higher attention coefficients ($e_{i,T} = 0.1707$) than CT players ($e_{i, CT} = 0.02924$). We reran our GAT model using only T and only CT players, and achieved an average log loss of 0.5609 for the T-only player model and 0.5714 for the CT-only player model. Thus, neither model performs well, indicating that both sides' information is still necessary to accurately predict round win probability. One explanation for the discrepancy in attention coefficients may be that if neither side performed any actions, the CT side would, by default, win the round, since the time would expire. Thus, the onus is on the T side to act.


Attention coefficients also varied across models. In Figure~\ref{fig:csgo_attn}, we show the difference between the combined GAT + State model and the GAT-only model for two different scenarios in the same CSGO round. In the ``pre-plant'' scenario, the T side has not yet planted the bomb, and the situation is 2 CT versus 3 T. In the ``defusing'' scenario, all T players have been eliminated, and the last CT player has to defuse the bomb to win the round. In the ``defusing scenario'', the GAT-only model estimates attention coefficients that are, on average, much lower for the remaining player than the GAT + State model. This is likely due to the GAT-only model unable to account for the little time remaining and the bomb is planted, which is information held in the state vector. 

Another difference is in the average attention coefficients of players in the upper-left (yellow box) portion of the map in the ``pre-plant'' scenario. In the GAT+State model, the T sided player ``Ethan'' has an attention coefficient that is 50\% higher than the coefficient estimated by the GAT-only model. Even though it is not visible due to CT attention coefficients generally being small, the CT player also has a higher attention coefficient relative to the other CT player alive. Both the T and CT player make up the majority of their team's total attention weight. This may indicate that the engagement that is taking place in the yellow-bounded box is important to the outcome of the round. The same set of CSGO domain experts from Section~\ref{sec:results} indicate that the engagement between Ethan and his opponent is important since it conveys important information to both teams, namely, if the CT player wins the engagement, they will know that there are no other players. If the T player (Ethan) wins the engagement, then the situation becomes three T players versus one CT player. In general, a coach or analyst may use attention coefficients to find high-value engagements in order to highlight key moments for game review. Furthermore, one could use attention coefficients as input to a player valuation system.

\subsection{Model Design Choices}
\subsubsection{Multi-Head Attention} \label{sec:attn_heads}
Velickovic~et~al.~\cite{VelickovicCCRLB18} suggest that employing multi-head learning may stabilize the self-attention learning process. We report the results of varying the number of attention heads on the NFL prediction task in Table~\ref{tab:multi-attention}. Using each attention head, we present the average MSE for 30 independent train/validation/test splits on the American football task. We conduct a paired t-test to compare each model to the single attention head model, and report the test statistic and p-values. Overall, we see that for the NFL prediction task, a single attention head is enough to achieve strong performance.

\begin{table}[]

\caption{Performance of GAT model by number of attention heads on the NFL prediction task. We see no significant statistical differences between using 2, 4 or 8 attention heads compared to a single attention head, which suggests that single attention head is enough for our NFL task.}

\centering
\begin{tabular}{@{}cccc@{}}
\toprule
\textbf{Attention Heads} & \textbf{MSE} & \textbf{t-statistic} & \textbf{p} \\ \midrule
1 & 30.04 & -- & -- \\
2 & 29.15 & 0.38 & 0.705 \\
4 & 35.00 & 2.01 & 0.054 \\
8 & 33.98 & -1.86 & 0.073 \\ \bottomrule
\end{tabular}

\label{tab:multi-attention}
\end{table}

\subsubsection{Keypoint-Based Features}
One of the benefits of graph-based models to predict sports outcomes is that agents no longer need to be ordered to ensure a permutation invariant output given the same sports game state. However, the question still remains over the necessity of keypoint-based features. For example, in our NFL task, we use each player's difference to the ball-carrier's speed and position as features. To test model performance without using keypoint-based features, we remove these features and rerun our GCN and GAT models on the NFL prediction task. Using a paired t-test, both the GAT ($t = -2.54$) and GCN ($t = -34.24$) model performance without keypoint-based features were significantly worse than the state-based model, thus, using keypoint-based features to represent player nodes is still important, particularly for American football. 

\subsection{Limitations and Future Work}
We see two major directions for future work -- model improvements and applications. One of the limitations of the current framework is that we use a single game state as input. While we improve upon the state of the art for single-state models, sequence-based models that consider multiple game states also perform well in sports prediction tasks. Yurko~et~al.~\cite{Yurko2020LSTM} showed how a recurrent neural network could consider sequences of game states represented as vectors to predict NFL outcomes. With this in mind, an interesting avenue to explore is if sequences of game states can be represented as graphs. Yan~et~al.~\cite{Yan18} propose representing sequences of human pose through a single graph, where different edges connect nodes not only within the same frame, but also temporally. Another limitation of our modeling approach was our graph pooling layer. Although we used global average pooling, which condenses our set of transformed player vectors into a single vector, one may consider different layer that downsample the graph to a coarser granularity. Finally, future work should explore how our proposed framework performs on tasks in other sports.

Our graph-based models can be used in a variety of applications in sports. One of the most applicable is player valuation. Similar to Xenopoulos~et~al.~\cite{Xenopoulos2020CSGOWPA}, our model which predicts CSGO round winners can be used to value player actions. Additionally, like Yurko~et~al.~\cite{Yurko2020LSTM}, our model to predict NFL rusher end yard line can be used to evaluate rusher performance. Aside from player valuation, we also see potential for our models to serve as the underpinning of visual analytics systems to analyze ``what if'' scenarios in sports. Similar to the work by Sha~et~al.~\cite{Sha18CHI}, we see valuable future work in developing and evaluating an interface that stakeholders, such as coaches and players, can use to investigate different positional setups and the corresponding change in predicted outcome. 

%% file: 06_conclusion.tex
\section{Conclusion} \label{sec:conclusion}
This paper introduces a sports-agnostic, graph-based representation of sports game states. Using this representation as input to graph convolutional networks, and evaluate our models both qualitatively and quantitatively. We show how our proposed networks provide statistically significant improvements over current state of the art approaches for prediction tasks in American football and esports (9\% and 20\% test set loss reduction, respectively). Furthermore, we assess our model's ability to represent well-known phenomena in sports, such as the advantageous effects of space in the NFL and high-impact events in CSGO. Besides improved prediction performance, one of the benefits of our modeling approach is that it enables the answering of ``what if'' questions and explicitly models interactions between players. We showcase our model's ability to assist sports analytics practitioners through examples in American football and esports. Furthermore, we visualize our model's estimated attention coefficients between players and show how the estimated coefficients reflect relevant game information.

%% file: 07_acks.tex
This work was partially supported by NSF awards: CCF-1533564, CNS-1544753, CNS-1730396,  CNS-1828576.